\documentclass[runningheads]{llncs}

\usepackage{algorithm}
\usepackage{algorithmic}
\usepackage{graphicx}
\usepackage{amsmath,amssymb}
\usepackage{color}

\begin{document}
\title{Super-Identity Convolutional Neural Network for Face Hallucination}

\titlerunning{Super-Identity Convolutional Neural Network for Face Hallucination}
%

\author{
Kaipeng Zhang\inst{1}  \and
Zhanpeng Zhang\inst{2}  \and
Chia-Wen Cheng\inst{1,3}  \and
Winston H. Hsu  \inst{1*}  \and
Yu Qiao\inst{4} \and
Wei Liu\inst{5} \and
Tong Zhang\inst{5}}

%
\authorrunning{K. Zhang et al.}
%

\institute{National Taiwan University, Taipei, Taiwan \\
\and
SenseTime Group Limited, China\\
\and
The University of Texas at Austin, USA \\
\and
Shenzhen Key Lab of Computer Vision and Pattern Recognition,
Shenzhen Institutes of Advanced Technology, CAS, Shenzhen, China \\
\and
Tencent AI Lab, China}
\maketitle              

\begin{abstract}
Face hallucination is a generative task to super-resolve the facial image with low resolution while human perception of face heavily relies on identity information. However, previous face hallucination approaches largely ignore facial identity recovery. This paper proposes Super-Identity Convolutional Neural Network (SICNN) to recover identity information for generating faces closed to the real identity. Specifically, we define a super-identity loss to measure the identity difference between a hallucinated face and its corresponding high-resolution face within the hypersphere identity metric space. However, directly using this loss will lead to a Dynamic Domain Divergence problem, which is caused by the large margin between the high-resolution domain and the hallucination domain. To overcome this challenge, we present a domain-integrated training approach by constructing a robust identity metric for faces from these two domains. Extensive experimental evaluations demonstrate that the proposed SICNN achieves superior visual quality over the state-of-the-art methods on a challenging task to super-resolve 12$\times$14 faces with an 8$\times$ upscaling factor. In addition, SICNN significantly improves the recognizability of ultra-low-resolution faces.

\end{abstract}
\section{Introduction}
\begin{figure}[t]
\begin{center}
 \includegraphics[width=1\linewidth]{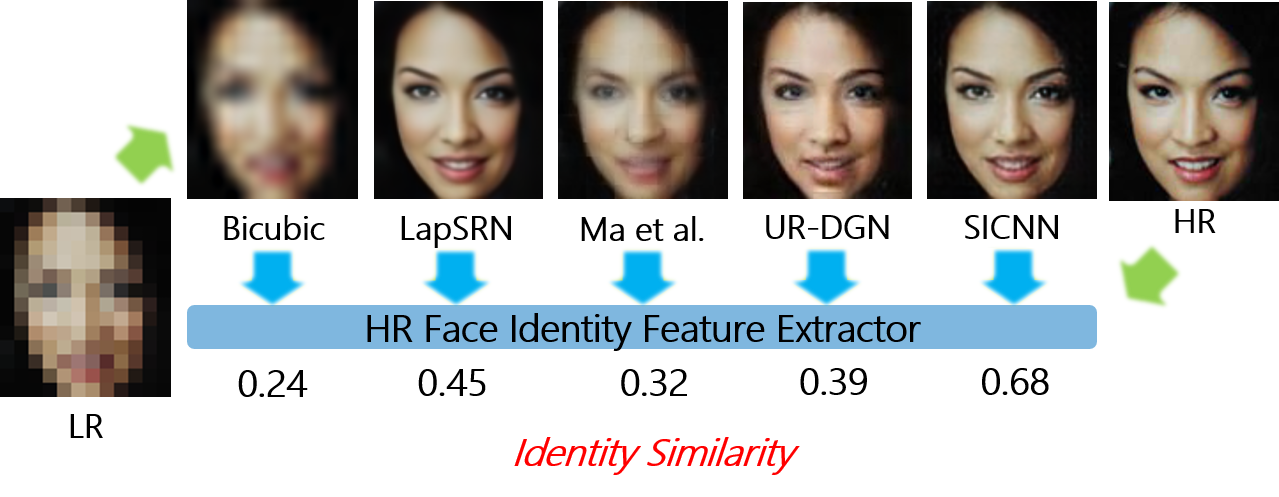}
\end{center}
   \caption{Comparison of face hallucination visual quality and the performance of identity recovery over different hallucination methods. The identity similarity is computed by the cosine similarity of the identity feature.}
\label{fig:long}
\label{fig:onecol}
\label{fig1}
\end{figure}

Face hallucination, which generates high-resolution (HR) facial images from low-resolution (LR) inputs, has attracted great interests in the past few years. However, most of existing works do not take the recovery of identity information into consideration such that they cannot generate faces closed to the real identity. Fig. \ref{fig1} shows some examples of hallucinated facial images generated by bicubic and several state-of-the-art methods. Though they generate clearer facial images than bicubic, the identity similarities are still low, which means that they cannot recover accurate identity-related facial details. On the other hand, human perception of face heavily relies on identity information \cite{idcode}. Pixel-level cues cannot fully account for the perception process of the brain. These facts suggest that recovering identity information may improve both the recognizability and performance of hallucination.

Motivated by the above observations, this paper proposes Super-Identity Convolutional Neural Network (SICNN) for identity-enhanced face hallucination. Different from previous methods, we additionally minimize the identity difference between the hallucinated face and its corresponding high-resolution face. To do so, (i) we introduce a robust identity metric space in the training process; (ii) we define a super-identity loss to measure the identity difference; (iii) we propose a novel training approach to efficiently utilize the super-identity loss. More details as follows:

For identity metric space, we use a hypersphere space \cite{sphere} as the identity metric space due to its state-of-the-art performance of facial identity representation. Specifically, our SICNN is composed of a face hallucination network cascaded with a recognition network to extract identity-related feature, and an Euclidean normalization operation to project the feature into the hypersphere space.

For loss function, perceptual loss \cite{johnson2016perceptual}, computed by feature Euclidean distance, can construct convincing HR images. Differently, in our work, we need to minimize the identity distance of face pairs in the metric space. Here, we modified the perceptual loss to the super-identity loss calculated by normalized Euclidean distance (equivalent to geodesic distance) between the hallucinated face and its corresponding high-resolution face in the hypersphere identity metric space. This also facilitates our analysis on the training process (see Sec. \ref{sec44}).

For training approach, using conventional training approaches to directly train the model with super-identity loss is difficult due to the large margin between the hallucination domain and the HR domain in the hypersphere identity metric space. This is critical during the early training stage when face hallucination network cannot predict high quality hallucinated face images. Moreover, the hallucination domain keeps changing during the hallucination network learning, which makes the training with super-identity loss unstable. We summarize this challenge as a \emph{dynamic domain divergence} problem. To overcome this problem, we propose a Domain Integrated Training algorithm that alternately updates the face recognition network and the hallucination network by minimizing the different loss in each iteration. In this alterative optimization, the hallucinated face and HR face will gradually move closer to each other in the hypersphere identity metric space while keep the discrimination of this metric space.

The main contributions of this paper are as summarized as follows:
\begin{itemize}
  \item We propose Super-identity Convolutional Neural Network (SICNN) for enhancing the identity information in face hallucination.
  \item We propose Domain-Integrated Training method to overcome the problem caused by dynamic domain divergence when training SICNN.
  \item Compared with existing state-of-the-art hallucination methods, the SICNN achieves superior visual quality and identity recognizability when super-resolving a facial image of size 12$\times$14 pixels with an $8\times$ upscaling factor.
\end{itemize}
\section{Related Works}
Single image super-resolution (SR) aims at recovering a HR image from a LR one. Face hallucination is a kind of class-specific image SR, which exploits the statistical properties of facial images. We classify face hallucination methods into two categories: classical approaches and deep learning approach.

\textbf{Classical Approaches}. Subspace-based and facial components-based methods are two main kinds of classical face hallucination approaches \cite{liu2007face,ma2010hallucinating,li2014face,tappen2012bayesian,yang2013structured,liuwei1,liuwei2,liuwei3}.

For subspace-based methods. Liu et al. \cite{liu2007face} employed a Principal Component Analysis (PCA) based global appearance model to hallucinate LR faces and a local non-parametric model to enhance the details. Ma et al. \cite{ma2010hallucinating} used multiple local exemplar patches sampled from aligned HR facial images to hallucinate LR faces. Li et al. \cite{li2014face} resolved to sparse representation on local face patches. These subspace-based methods require precisely aligned reference HR and LR facial images with the same pose and facial expression.

Facial components based methods super-resolve facial parts rather than entire faces to address various poses and expressions. Tappen et al. \cite{tappen2012bayesian} used SIFT flow to align LR images, and then deformed the reference HR images. However, the global structure is not preserved due to using local mapping. Yang et al. \cite{yang2013structured} presented a structured face hallucination method which can maintain the facial structure. However, it relies on accurate facial landmarks.

\textbf{Deep Learning Approaches}. Recently, deep convolutional neural networks (DCNNs) achieve remarkable progresses in a variety of face analysis tasks, such as face recognition \cite{cosface,centerloss,sphere}, face detection \cite{mtcnn,iccv17}, facial attribute recognition \cite{cvprw16,celebrity,emotiw17,emotiw18}. Zhou et al. \cite{zhou2015learning} proposed a bichannel CNN to hallucinate blurry facial images in the wild. For un-aligned faces, Zhu et al. \cite{zhu2016deep} proposed to jointly learn face hallucination and facial dense spatial correspondence field estimation. The approach of \cite{urdgn} is a GAN-based method to generate realistic facial images. These works ignore the identity information recovery that is important for recognizability and hallucination quality. Johnson et al. \cite{johnson2016perceptual} and Bruna et al. \cite{bruna2015super} relied on perceptual loss function closer to perceptual similarity to recover visually more convincing HR images for general image SR. In this paper we modified the perceptual loss to facilitate identity hypersphere space and propose a novel training approach to overcome the challenging while using the loss.
\section{Super-Identity CNN}
In this section, we will first describe the architecture of our face hallucination network. Then we will introduce the proposed super-resolution loss and super-identity loss for identity recovery. After that, we will analyze the challenge, dynamic domain divergence problem, in super-identity training. At the last, we introduce the proposed domain-integrated training algorithm to overcome this challenge.
\subsection{Face Hallucination Network Architecture}\label{sec40}
\begin{figure}[t]
\begin{center}
 \includegraphics[width=1\linewidth]{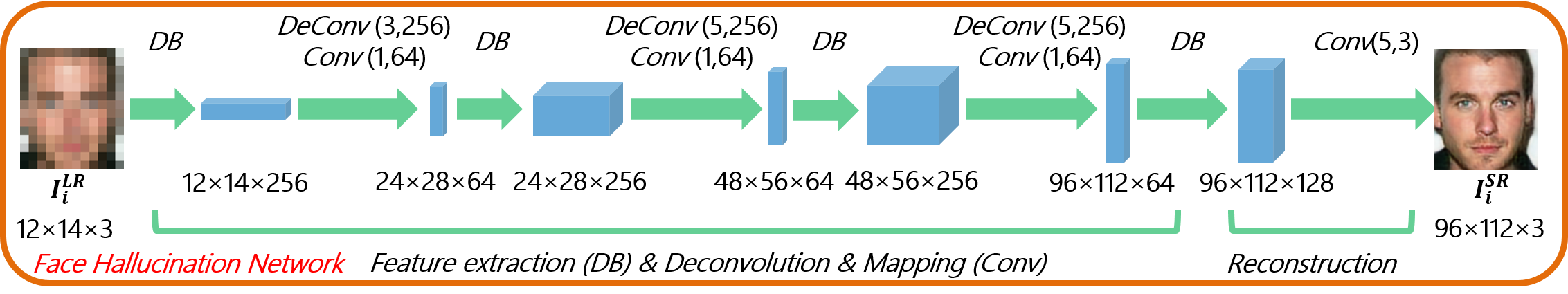}
   (a) Network architecture of hallucination model ($CNN_{H}$)
   \includegraphics[width=1\linewidth]{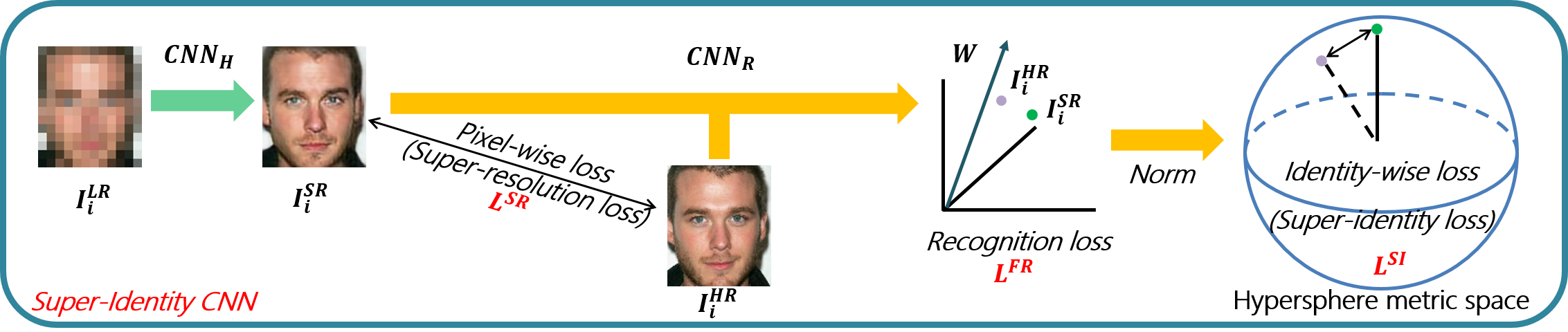}
   (b) Illustration of the proposed super-identity CNN
\end{center}
   \caption{Framework of our approach. (a) The network architecture of our hallucination network ($CNN_{H}$). DB denotes dense block \cite{densenet}. (b) Illustration of our super-identity CNN. It uses super-resolution loss ($L^{SR}$), super-identity loss ($L^{SI}$), and recognition loss ($L^{FR}$) with domain-integrated training. Norm denotes Euclidean normalization, and $CNN_{R}$ denotes the recognition network.}
\label{fig:long}
\label{fig:onecol}
\label{fig2}
\end{figure}

As shown in Fig. \ref{fig2} (a), the face hallucination network can be decomposed into feature extraction, deconvolution, mapping, and reconstruction.

We use dense block \cite{densenet} to extract semantic features from LR inputs.
More specifically, in the dense block, we set the growth rate to 32 and the kernel size to 3$\times$3. Deconvolution layer consists of learnable upscaling filters to enlarge the resolutions of input features. Mapping is implemented by a convolutional layer to reduce the dimension of features to reduce computational cost. Reconstruction also exploits a convolutional layer to predict HR images from semantic features.

Here, we denote a convolutional layer as $Conv(s,c)$ and a deconvolutional layer as $DeConv(s,c)$, where the variables $s$ and $c$ represent the filter size and the number of channels, respectively. In addition, PReLU \cite{prelu} activation function achieves promising performance in CNN-based super-resolution \cite{fsrcnn} and we use it after each layer except the reconstruction stage.
\subsection{Super-Resolution Loss}\label{sec41}
We use the pixel-wise Euclidean loss, called super-resolution loss, to constrain the overall visual appearance. For LR face input $I^{LR}_i$, we penalize the pixel-wise Euclidean distance between the hallucinated face and its corresponding HR face:
\begin{equation}
L^{SR}(I^{LR}_i,I^{HR}_i)=\left \| CNN_{H}(I^{LR}_i)- I^{HR}_i\right \|_{2}^{2}
\label{eq1},
\end{equation}
where $I^{LR}_i$ and $I^{HR}_i$ are the $i$-th LR and HR facial image pair in the training data respectively, and $CNN_{H}(I^{LR}_i)$ represents the output of hallucination network with input $I^{LR}_i$. For better understanding, we also denote $CNN_{H}(I^{LR}_i)$ as $I^{SR}_i$ in the following text.

\subsection{Hypersphere Identity Metric Space}\label{sec42}
Super-resolution loss can constrain pixel-level appearance. And we further use a constrain on the identity level. To measure the identity level difference, the first step is to find a robust identity metric space. Here we employ the hypersphere space \cite{sphere} due to its state-of-the-art performance on identity representation. As shown in Fig. \ref{fig2} (b), our hallucination network is cascaded with a face recognition network (i.e. $CNN_{R}$) and an Euclidean normalization operation that projects faces to the constructed hypersphere identity metric space.

$CNN_{R}$ is a Resnet-like \cite{resnet} CNN (see Tab. \ref{tablenn}). It is trained by A-Softmax loss function \cite{sphere} which encourages the CNN to learn discriminate identity features (i.e. maximizing inter-class distance and minimizing intra-class distance) by an angular margin. In this paper, we denote this loss function as the recognition loss $L^{FR}$. For a face input $I_{i}$ belonging to the $y_{i}$-th identity. The face recognition loss is represented as:
\begin{equation}
L^{FR}(I_{i})=-\log(\frac{e^{\left \| CNN_{R}(I_{i}) \right \| \varphi(m\Theta_{y_{i}})}}{e^{\left \| CNN_{R}(I_{i}) \right \| \varphi(m\Theta_{y_{i}})}+\sum _{j\neq y_{i}}e^{\left \| CNN_{R}(I_{i}) \right \| \varphi(\Theta_{j})}}),
\label{eqex}
\end{equation}
where the $\Theta_{y_{i}}$ denotes the learned angle for identity $y_{i}$, $\varphi(\Theta_{y_{i}})$ is a monotonically decreasing function generalized from $\cos(\Theta_{y_{i}})$, and $m$ is the hyper parameter of angular margin constrain. More details can be found in Sphereface \cite{sphere}.

\begin{table}[t]
\begin{center}
\begin{tabular}{|l|c|c|}
\hline
Layer Name&Output Size&Structure\\
\hline
Input&96$\times$112&-\\
\hline
Conv1a&94$\times$110&3$\times$3, 64, pad 0\\
\hline
Conv1b&92$\times$108&3$\times$3, 64, pad 0\\
\hline
Avepool1&46$\times$54&3$\times$3, stride 2\\
\hline
Residual\_block1
&
46$\times$54
&
$
\begin{bmatrix}
3\times3, 64\\
3\times3, 64\\
\end{bmatrix}
\times1
$
\\
\hline
Conv2&44$\times$52&3$\times$3, 128, pad 0\\
\hline
Avepool2&22$\times$26&3$\times$3, stride 2\\
\hline
Residual\_block2
&
22$\times$26
&
$
\begin{bmatrix}
3\times3, 128\\
3\times3, 128\\
\end{bmatrix}
\times2
$
\\
\hline
Conv3&20$\times$24&3$\times$3, 256, pad 0\\
\hline
Avepool3&10$\times$12&3$\times$3, stride 2\\
\hline
Residual\_block3
&
10$\times$12
&
$
\begin{bmatrix}
3\times3, 256\\
3\times3, 256\\
\end{bmatrix}
\times5
$
\\
\hline
Conv4&8$\times$10&3$\times$3, 512, pad 0\\
\hline
Avepool4&4$\times$5&3$\times$3, stride 2\\
\hline
Residual\_block4
&
4$\times$5
&
$
\begin{bmatrix}
3\times3, 512\\
3\times3, 512\\
\end{bmatrix}
\times3
$
\\
\hline
FC1&512&4$\times$5, 512\\
\hline
\end{tabular}
\end{center}
\caption{The architecture for our face recognition CNN ($CNN_{R}$). It follows the residual block structure \cite{resnet}. We use PReLU \cite{prelu} activation function after each convolution layer. The output of FC1 is the identity representation.}
\label{tablenn}
\end{table}
\subsection{Super-Identity Loss}\label{sec43}
To impose the identity information in the training process, one choice is to use a loss computed by features Euclidean distance between face pairs, such as perceptual loss \cite{johnson2016perceptual}. However, in this paper, since our goal is to minimize identity distance in hypersphere metric space, the original perceptual loss, computed by L2 distance is not the best choice in our task. Therefore, we propose a modified perceptual loss, called Super-Identity (SI) loss, to compute the normalized Euclidean distance (equivalent to geodesic distance). This modification makes the loss directly related to identity in hypersphere space and facilitate our investigation in Sec. \ref{sec44}.

For a LR face input $I^{LR}_i$, we penalize the normalized Euclidean distance between the hallucinated face and its corresponding HR face in the constructed hypersphere identity metric space:
\begin{equation}
L^{SI}(I^{LR}_i,I^{HR}_i)=\left\|\widehat{CNN_{R}(I^{SR}_i)}-\widehat{CNN_{R}(I^{HR}_i)}\right \|_{2}^{2}
\label{eq3}
\end{equation}
where $CNN_{R}(I^{SR}_i)$ and $CNN_{R}(I^{HR}_i)$ are the identity features extracted from face recognition model ($CNN_{R}$) for facial images $I^{SR}_i$ and $I^{HR}_i$, respectively. $\widehat{CNN_{R}(I^{SR}_i)}=\frac{CNN_{R}(I^{SR}_i)}{\left\|CNN_{R}(I^{SR}_i)\right \|_{2}}$ is the identity representation projected to the unit hypersphere.

In addition to $L^{SI}$, we want to have some discussions about perceptual loss beyond our work. In general, the perceptual loss is computed by L2 distance. However, in most CNNs, inner-product operation is used in fully-connected and convolutional layers. These outputs are related to the feature's norm, weight's norm and the angular between them. Therefore, for different tasks and different metric space (e.g. \cite{spherenet,cosinenorm,dml}), some modifications about computational metric space of perceptual loss are necessary ($L^{SI}$ is one of the cases).
\subsection{Challenges of Training with Super-Identity Loss}\label{sec44}
Super-identity loss imposes an identity level constrain. We examine different training methods as follows:

\noindent \textbf{Baseline training approach I.} A straightforward way to train our framework is jointly using the $L^{SR}$, $L^{SI}$ and $L^{FR}$ to train both $CNN_{H}$ and $CNN_{R}$ from scratch. The optimization objective can be represented as:
\begin{equation}
\underset{\theta_{CNN_{H}}\theta_{CNN_{R}}}{\min}\frac{1}{n}\sum_{i=1}^{n}L^{SR}(I^{LR}_i,I^{HR}_i) + \alpha L^{SI}(I^{LR}_i,I^{HR}_i) + \beta L^{FR}(I^{SR}_i,I^{HR}_i),
\label{eqex1}
\end{equation}
where $\alpha$ and $\beta$ denotes the loss weight of the $L^{SI}$ and $L^{FR}$ respectively, $\theta_{CNN_{H}}$ and $\theta_{CNN_{R}}$ denotes the learnable parameters.
\begin{figure}
\begin{center}
 \includegraphics[width=1\linewidth]{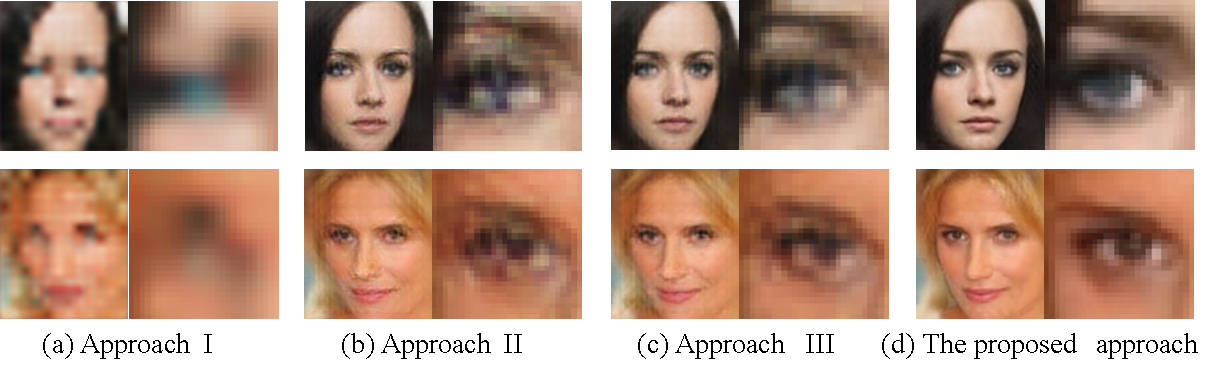}
\end{center}
   \caption{Face hallucination examples produced by $CNN_{H}$ trained by different training approaches. These four columns of results are produced by baseline training approach I, II, III and the proposed domain-integrated training approach respectively. It is clear that our approach achieves the best result while other results are noisy. This figure is best viewed in color. Please \emph{zoom in} for better comparison.}
\label{fig:long}
\label{fig:onecol}
\label{fig4}
\end{figure}

\noindent \textbf{Observation I.} This training approach generates artifacts (see Fig. \ref{fig4}, first column) and the loss is too difficult to converge. The reasons may come from: (1) In the early training stage, the hallucinated faces are quite different from HR faces, so the $CNN_{R}$ is too difficult to be optimized from scratch. (2) The objective of $L^{FR}$ (i.e. minimizing the intra-class variance) is different from the objective of $L^{SI}$ and $L^{SR}$ loss (minimizing the pair-wise distance), which is disadvantageous to $CNN_{R}$ and $CNN_{H}$ learning. So, we cannot use the $L^{SI}$ in $CNN_{R}$ learning and also cannot use the $L^{FR}$ in $CNN_{H}$ learning.

\noindent \textbf{Baseline training approach II.} To solve above problems, one possible training approach used in perceptual loss \cite{johnson2016perceptual} can be used. In particular, we train a $CNN_{R}$ using HR faces and then jointly use the $L^{SR}$ and the $L^{SI}$ to train the $CNN_{H}$. The joint objective of $L^{SI}$ and $L^{SR}$ can be represented as:
\begin{equation}
\underset{\theta_{CNN_{H}}}{\min}\frac{1}{n}\sum_{i=1}^{n}L^{SR}(I^{LR}_i,I^{HR}_i) + \alpha L^{SI}(I^{LR}_i,I^{HR}_i),
\label{eq4}
\end{equation}

\begin{figure}
\begin{center}
 \includegraphics[width=1\linewidth]{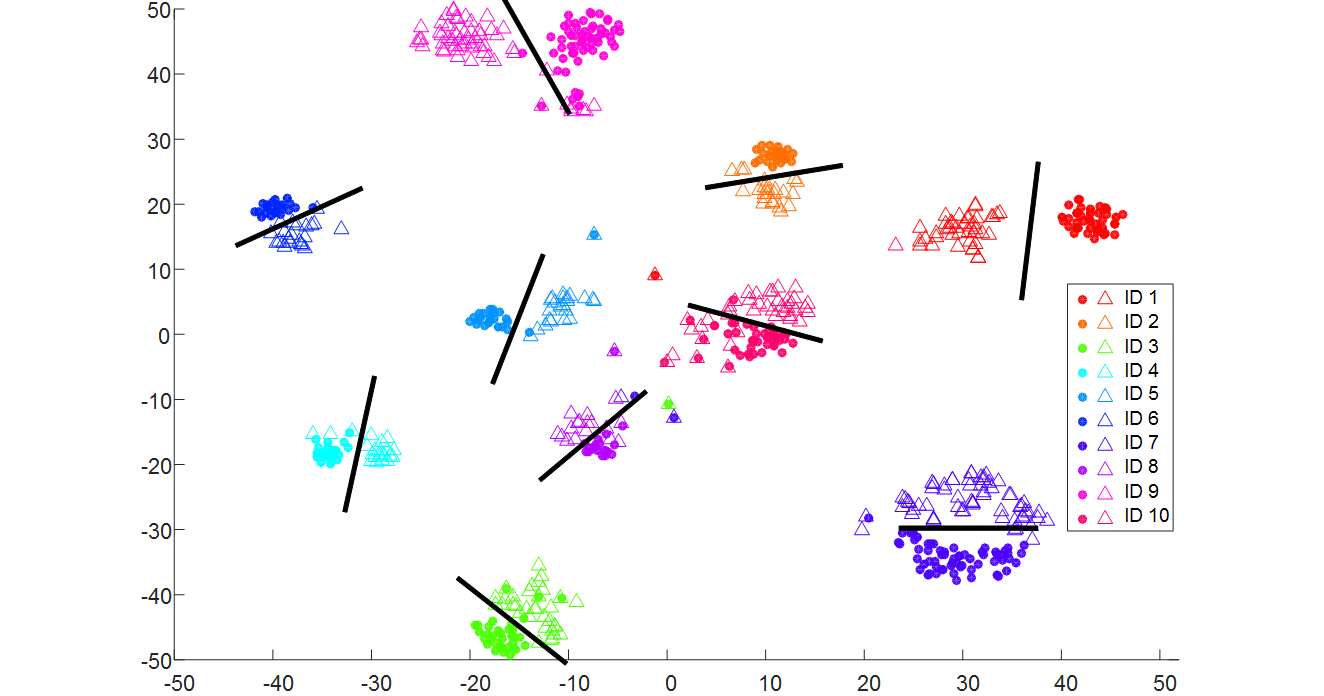}
\end{center}
   \caption{The distribution of identity features (after Euclidean normalization) from hallucination domain (triangle) and HR domain (dot). These identities are randomly selected from the training set. Different colors denote different identities. We use t-SNE \cite{tsne} to reduce the dimensions for better understanding. We can observe that there is a large gap between above two domains in the identity metric space.}
\label{fig:long}
\label{fig:onecol}
\label{fig5}
\end{figure}

\noindent \textbf{Observation II.} We have two observations while using this training approach: (1) The $L^{SI}$ is difficult to converge. (2) The visual results are noisy (see Fig. \ref{fig4}, second column). To investigate these challenges, we first visualized the learned identity features (after Euclidean normalization, as shown in Fig. \ref{fig5}) and found that there exists a large margin between the hallucination domain and the HR domain. We formulate this challenge as domain divergence problem. It specifies the failure of the $CNN_{R}$, trained by HR faces, to project faces from hallucination domains to a measurable hypersphere identity metric space. In other words, this face recognition model cannot extract effective identity representation for hallucinated faces. This makes the $L^{SI}$ very difficult to converge and easily get stuck in local minima (i.e. occur many noises in hallucination results).

\noindent \textbf{Baseline training approach III.} To overcome the domain divergence challenge, a straightforward alternately training strategy can be used. In particular, we first trained a $CNN_{H}$ only using the $L^{SR}$. Then we trained a $CNN_{R}$ using hallucinated faces and HR faces. Finally, we finetune the $CNN_{R}$ jointly using the $L^{SR}$ and the $L^{SI}$ following baseline training approach II.

\noindent \textbf{Observation III.} Although this alternately training strategy seems able to overcome the domain divergence problem, it still produces artifacts (as shown in Fig. \ref{fig4}, third column). The reason is that the hallucination domain keeps changing when the $CNN_{H}$ is being updated. If the hallucination domain has changed, the face recognition model cannot extract effective and measurable identity representation of hallucinated faces anymore.

In short, above observations can be concluded into a dynamic domain divergence problem as following: a large margin exists between the hallucination domain and HR domain and the hallucination domain keeps changing if the hallucination model keeps learning.

\begin{algorithm}[t]
  \caption{Mini-batch SGD based domain-integrated training approach}
  \label{alg1}
  {\bf Input:}
  Face recognition model $CNN_{R}$ trained by HR facial images, face hallucination model $CNN_{H}$ trained by $L^{SR}$, minibatch size $N$, LR and HR facial image pairs $\left \{I^{LR}_i,I^{HR}_i \right \}$.

  {\bf Output:}
  SICNN.
  \begin{algorithmic}[1]
  \WHILE{not converge}
  \STATE Choose one minibatch of $N$ LR and HR image pairs $\left \{I^{LR}_i,I^{HR}_i \right \}$, $i=1,...,N$.
  \STATE Generate one minibatch of $N$ hallucinated facial images $I^{SR}_i$ from $I^{LR}_i$, $i=1,...,N$, where $I^{SR}_i=CNN_{H}(I^{LR}_i)$.
  \STATE Update the recognition model $CNN_{R}$ by descending its stochastic gradient:
  \begin{center}
    $\bigtriangledown_{\theta_{CNN_{R}}}\frac{1}{N}\sum_{i=1}^{N}L^{FR}(\left \{ I^{SR}_i,I^{HR}_i \right \})$
  \end{center}
  \STATE Update the hallucination model $CNN_{H}$ by descending its stochastic gradient:
  \begin{center}
    $\bigtriangledown_{\theta_{CNN_{H}}}\frac{1}{N}\sum_{i=1}^{N} L^{SR}(I^{LR}_i,I^{HR}_i)+\alpha L^{SI}(I^{LR}_i,I^{HR}_i)$
  \end{center}
  \ENDWHILE
  \end{algorithmic}
\end{algorithm}
\subsection{Domain-Integrated Training Algorithm}\label{sec45}
To overcome the dynamic domain divergence problem, we propose a new training procedure. From above the above observations, we see that alternately training strategy (Baseline Training Approach III) can alleviate the dynamic domain divergence problem. We further propose to do this alternately training in each iteration.

More specifically, we first train a $CNN_{R}$ using HR facial images and a $CNN_{H}$ using the $L^{SR}$. Then, we propose to use domain-integrated training approach (Algorithm \ref{alg1}) to finetune $CNN_{R}$ and $CNN_{H}$ alternately in each iteration.

In particular, in each iteration, we first update the $CNN_{R}$ using the recognition loss, which allows the $CNN_{R}$ to perform accurate identity representation in this mini-batch of faces from different domains. Then, we jointly use the $L^{SR}$ and the $L^{SI}$ to update the $CNN_{H}$. This training approach can encourage the $CNN_{R}$ to construct a robust mapping from faces to the measurable hypersphere identity metric space in each iteration for $L^{SI}$ optimization whatever the $CNN_{H}$ is changing. The alternative optimization process is conducted until converged. Some hallucination examples are shown in Fig. \ref{fig4}, fourth column, where we can observe a much better visual result with this training approach.
\subsection{Comparison to Adversarial Training}\label{sec46}
Domain-Integrated (DI) training and adversarial training \cite{gan} can be related to their alternative learning strategy. But they are quite different in several aspects as follows:

(1) Generally speaking, DI training is essentially a cooperative process in which $CNN_{H}$ collaborates with $CNN_{R}$ to minimize the identity difference. The learning objective is the same in each sub-iteration. However, in adversarial training, generator and discriminator compete against each other to improve the performance. The learning objective is alternatively challenging during two models learning.

(2) The loss functions and optimization style are different. In DI training, we minimize $L^{FR}$ in $CNN_{R}$ constructing a marginal identity metric space and then minimize $L^{SI}$ for $CNN_{H}$ reducing pair-wise identity difference. Differently, in adversarial training, the classification loss is minimized for discriminator learning and maximized for generator learning.

\section{Experiments}
In this section, we will first describe the training and testing details. Then we perform an ablation study to evaluate the effectiveness of the proposed Super-Identity loss and Domain-Integrated training. Further, we evaluate our proposed method with other state-of-the-art methods. After that, we evaluate our method on the higher input size. At the last, we evaluate the benefit of our method for low-resolution face recognition.
\subsection{Training Details}
\noindent \textbf{Training data.} For a fair comparison with other state-of-the-art methods, we do face alignment in facial images. In particular, we use similarity transformation based on five landmarks detected by MTCNN \cite{mtcnn}.  \emph{We have removed the images and identities overlap between training and testing}.

For face recognition training, we use web-collected facial images including CASIA-WebFace \cite{casia}, CACD2000 \cite{cacd}, CelebA \cite{celebrity}, VGG Faces \cite{vggface} as \emph{Set A}. It roughly goes to 1.5M images of 17,680 unique persons.

For face hallucination training, we select 1.1M HR facial images (larger than 96$\times$112 pixels) from the same 1.5M images as \emph{Set B}.

\noindent \textbf{Training details.} For recognition model training, we use \emph{Set A} with the batch size of 512 and $m$ (angular margin constrain in Eq. \ref{eqex}) of 4. The learning rate is started from 0.1 and divided by 10 at the 20K, 30K iterations. The training process is finished at 35K iterations.

For hallucination model training, we use \emph{Set B} with the batch size of 128. The learning rate is started from 0.02 and divided by 10 at the 30K, 60K iterations. A complete training is finished at 80K iterations.

For domain-integrated training, we use \emph{Set B} with the batch size of 128 for $CNN_{H}$ and 256 for $CNN_{R}$. The learning rate is started from 0.01 and divided by 10 at the 6K iterations. A complete training is finished at 9K iterations.
\subsection{Testing Details}
\noindent \textbf{Testing data.} We randomly select 1,000 identities with 10,000 HR facial images (larger than 96$\times$112 pixels) from UMD-Face \cite{umdface} dataset as \emph{Set C}. The dataset is used for face hallucination and identity recovery evaluation.


\noindent \textbf{Evaluation protocols.} In this section, we perform three kinds of evaluations: (1) Visual quality. (2) Identity recovery. (3) Identity recognizability. For visual quality evaluation, we report several visual examples  results on \emph{Set C}.

For identity recovery, we evaluate the performance of recovering identity information while super-resolving faces. In particular, we use the $CNN_{R}$ trained by \emph{Set A} as identity features extractor. And the identity features are taken from the output of the first fully connected layer. Then we compute the identity similarity (i.e. cosine similarity) between the hallucinated face and its corresponding HR faces on \emph{Set C}. The average similarities over the testing set are reported.

For identity recognizability, we evaluate the recognizability of hallucinated faces. In particular, we first downsample \emph{Set A} to 12$\times$14 pixels as \emph{Set A - LR}. Then we use different methods to super-resolve \emph{Set A - LR} to 96$\times$112 pixels as different \emph{Set A - SR}. At last, we use the \emph{Set A - SR} to train different $CNN_{R}$ and evaluate them on LFW \cite{lfw} and YTF \cite{ytf}.
\begin{figure}
\begin{center}
\begin{minipage}[t]{0.118\linewidth}
\centering
\includegraphics[width=1\linewidth]{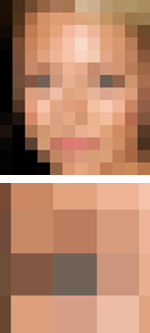}

LR
\end{minipage}%
\begin{minipage}[t]{0.118\linewidth}
\centering
\includegraphics[width=1\linewidth]{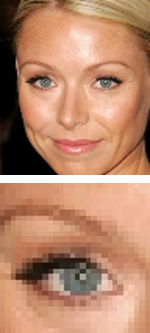}

HR
\end{minipage}
\begin{minipage}[t]{0.118\linewidth}
\centering
\includegraphics[width=1\linewidth]{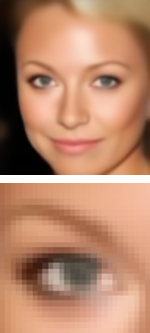}

$\alpha=0$
\end{minipage}
\begin{minipage}[t]{0.118\linewidth}
\centering
\includegraphics[width=1\linewidth]{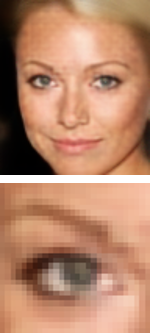}

$\alpha=2$
\end{minipage}
\begin{minipage}[t]{0.118\linewidth}
\centering
\includegraphics[width=1\linewidth]{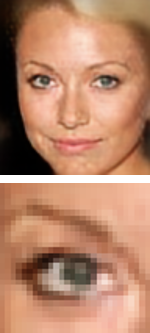}

$\alpha=4$
\end{minipage}
\begin{minipage}[t]{0.118\linewidth}
\centering
\includegraphics[width=1\linewidth]{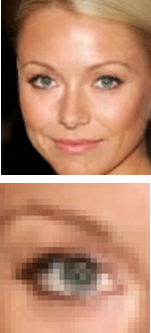}

$\alpha=8$
\end{minipage}
\begin{minipage}[t]{0.118\linewidth}
\centering
\includegraphics[width=1\linewidth]{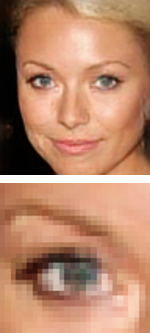}

$\alpha=16$
\end{minipage}
\begin{minipage}[t]{0.118\linewidth}
\centering
\includegraphics[width=1\linewidth]{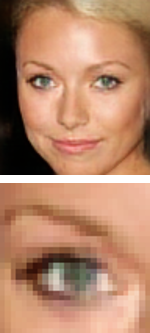}

$\alpha=32$
\end{minipage}
\end{center}
   \caption{Face hallucination examples generated by models trained with different loss weight $\alpha$. It is clear that choosing larger $\alpha$ can make the facial images sharper with more details. Please \emph{zoom in} for better comparison.}
\label{fig:long}
\label{fig:onecol}
\label{fig6}
\end{figure}
\subsection{Ablation Experiment}\label{sec52}
\textbf{Loss weight.} The hyper parameter $\alpha$ (see Algorithm 1) dominates the identity recovery. To verify the effectiveness of the proposed Super-Identity loss, we vary $\alpha$ from 0 (i.e. only use super-resolution loss) to 32 to learn different models. From Tab. \ref{tab1} and Fig. \ref{fig6}, we observe that larger $\alpha$ make the facial images sharper with more details and brings the better performance of identity recovery and recognizability. But too large $\alpha$ also makes the texture look slightly unnatural. And, since the performances of identity recovery and identity recognizability are stable when $\alpha$ is larger than 8, we fix $\alpha$ to 8 in other experiments.

\begin{table}
\begin{center}
\begin{tabular}{|l|c|c|c|c|c|c|}
\hline
$\alpha$ & 0 & 2 & 4 & 8 & 16 & 32 \\
\hline
Identity Similarity & 0.4418 & 0.5134& 0.5639& 0.5978& 0.6041& 0.6101\\
\hline
LFW Accuracy & 97.61\% & 97.88\%& 98.05\%& 98.25\%& 98.23\%& 98.16\%\\
\hline
YTF Accurarcy & 93.20\% & 93.48\%& 93.56\%& 93.82\%& 93.84\%& 93.76\%\\
\hline
\end{tabular}
\end{center}
\caption{Quantitative comparison of different $\alpha$ on identity recovery and identity recognizability evaluation. Larger $\alpha$ brings better performance and it is stable when $\alpha$ is larger than 8.}
\label{tab1}
\end{table}

\textbf{Training approach.} We evaluate different training approaches introduced in Sec. \ref{sec44} and Sec. \ref{sec45}. Some visual results are shown in Fig. \ref{fig4}. We can see that Domain-Integrated training achieves the best visual results. Besides, from Tab. \ref{tab2}, Domain-Integrated training also achieves the best performance of identity recovery and identity recognizability.

\begin{table}
\begin{center}
\begin{tabular}{|l|c|c|c|c|}
\hline
Training Approach &  I & II & III & Domain-Integrated Training \\
\hline
Identity Similarity & 0.3875 & 0.4829& 0.5132& \bf{0.5978}\\
\hline
LFW Accuracy & 97.16\% & 97.46\%& 97.58\%& \bf{98.25\%}\\
\hline
YTF Accurarcy & 92.98\% & 93.32\%& 93.34\%& \bf{93.84\%}\%\\
\hline
\end{tabular}
\end{center}
\caption{Quantitative comparison of different training approaches on identity recovery and identity recognizability evaluation. The results demonstrate the superiority of our proposed domain-integrated training.}
\label{tab2}
\end{table}
\subsection{Evaluation on Face Hallucination}\label{sec53}
We compare SICNN with other state-of-the-art methods and bicubic interpolation on \emph{Set C} for face hallucination. In particular, we follow EnhanceNet \cite{enhancenet} training another UR-DGN, called UR-DGN*, with additional perceptual loss computed in end of the second and the last ResBlock in $CNN_{R}$. \emph{All methods are re-trained in same training set - Set B.}

\begin{figure}[H]
\begin{center}
\begin{minipage}[t]{0.118\linewidth}
\centering
\includegraphics[width=1\linewidth]{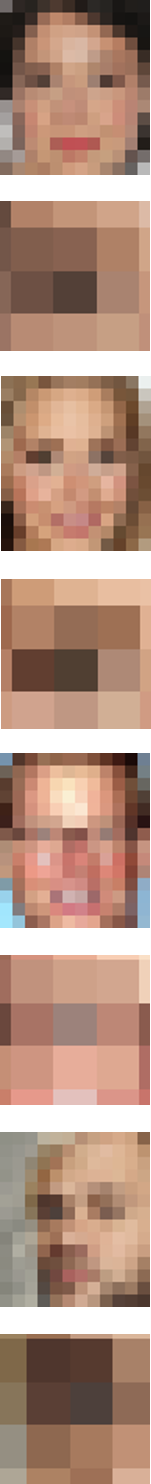}
LR
\end{minipage}%
\begin{minipage}[t]{0.118\linewidth}
\centering
\includegraphics[width=1\linewidth]{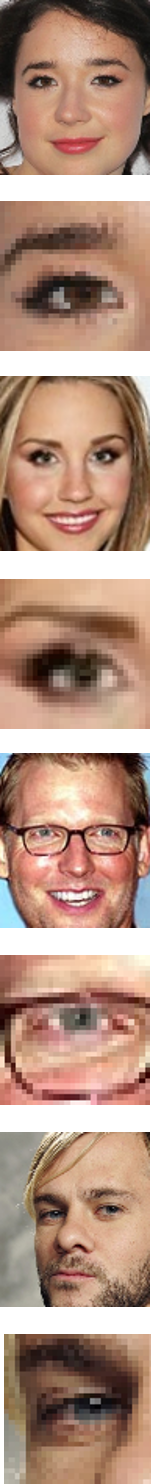}
HR
\end{minipage}
\begin{minipage}[t]{0.118\linewidth}
\centering
\includegraphics[width=1\linewidth]{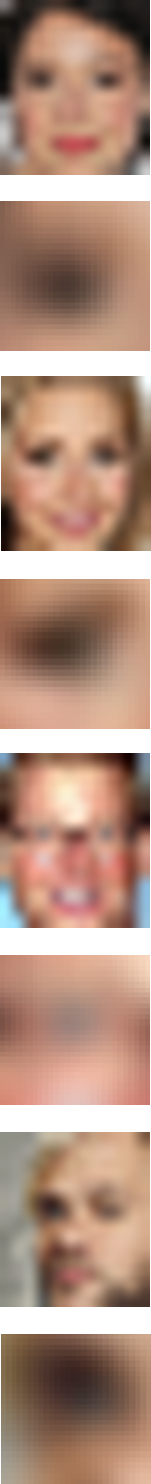}
Bicubic
\end{minipage}
\begin{minipage}[t]{0.118\linewidth}
\centering
\includegraphics[width=1\linewidth]{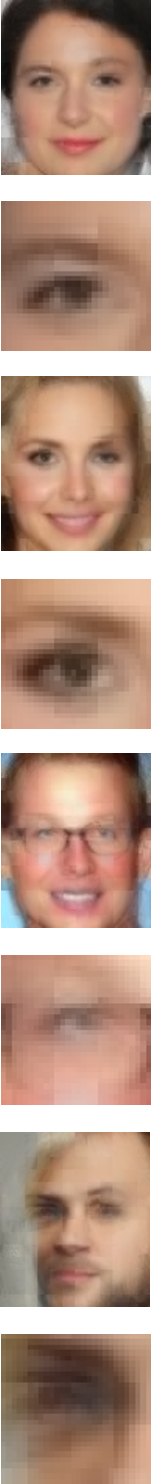}
Ma et al.
\end{minipage}
\begin{minipage}[t]{0.118\linewidth}
\centering
\includegraphics[width=1\linewidth]{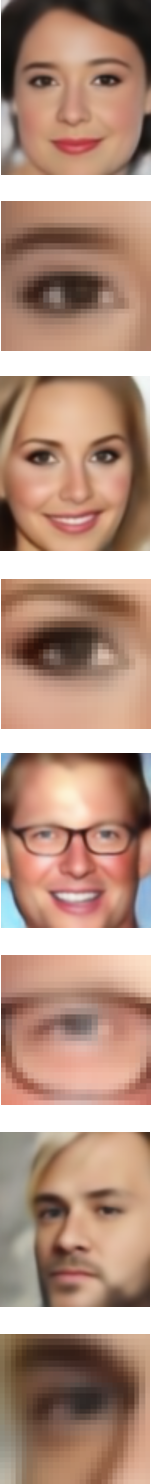}
LapSRN
\end{minipage}
\begin{minipage}[t]{0.118\linewidth}
\centering
\includegraphics[width=1\linewidth]{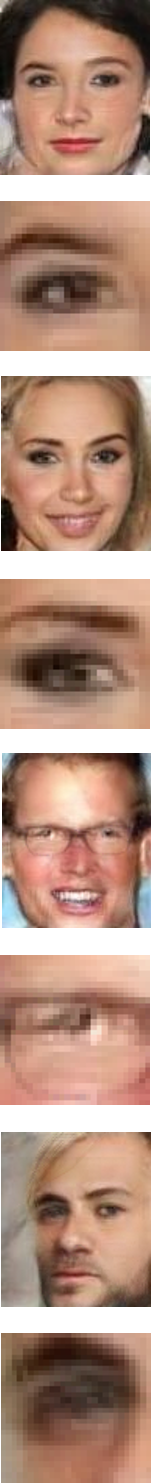}
URDGN
\end{minipage}
\begin{minipage}[t]{0.118\linewidth}
\centering
\includegraphics[width=1\linewidth]{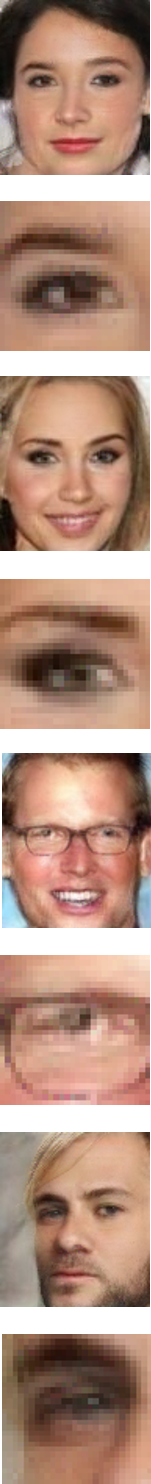}
URDGN*
\end{minipage}
\begin{minipage}[t]{0.118\linewidth}
\centering
\includegraphics[width=1\linewidth]{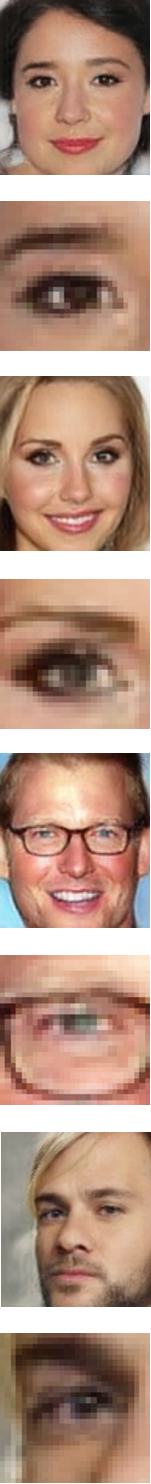}
SICNN
\end{minipage}
\label{fig:long}
\label{fig:onecol}
\end{center}
   \caption{Comparison with the state-of-the-art methods on hallucination test dataset. It is clear that our method achieves the best hallucination visual quality. Please \emph{zoom in} for better comparison. More visual results are included in our supplementary material.}
\label{fig:long}
\label{fig:onecol}
\label{fig8}
\end{figure}

Some visual examples are shown in Fig. \ref{fig8}. More visual results are included in our supplementary material.
We also report the results of average Peak Signal-to-Noise Ratio (PSNR) and Structural Similarity (SSIM) in Tab. \ref{tab3}.
But as the claim of other works \cite{johnson2016perceptual,enhancenet,ledig2016photo}, PSNR and SSIM results are useless for sematic super-resolution evaluation while visual quality and recognizability are more valuable.

From the visual results, it is clear that our method achieves the best results over other methods. We analyze the results as follows:

(1) For Ma \emph{et al.}'s method, exemplar patches based, the results are over-smooth and suffer from obvious blocking for such low low-resolution input with large up-sampling scale.

(2) For LapSRN \cite{lapsrn}, since it is based on L2 pixel-wise loss, it makes the hallucinated faces over-smooth.

(3) For UR-DGN \cite{urdgn}, it jointly uses pixel-wise Euclidean loss and adversarial loss to generate a realistic facial image closest to the average of all potential images. Thus, though the generated facial images look realistic, they are quite different from the original HR images.

(4) For UR-DGN*, it uses an additional loss - perceptual loss computed in our $CNN_{R}$ as the pair-wise semantic loss for identity recovery. Though this pixels-wise loss + adversarial loss + perceptual loss is the state-of-the-art super-resolution training approach (i.e. EnhancementNet\cite{enhancenet}). It still achieves inferior results than ours.

\begin{table}[t]
\begin{center}
\begin{tabular}{|l|c|c|c|c|c|c|c|c|}
\hline
Method & Bicubic &  Ma \emph{et al.}&LapSRN  & UR-DGN & UR-DGN* & SICNN\\
\hline
PSNR (db) & 23.1323 & 23.8606 & 26.1451 & 24.1857 & 25.2859 & 26.8945  \\
\hline
SSIM & 0.6093 & 0.6571 & 0.7417 & 0.6764 & 0.7224 & 0.7689  \\
\hline
\end{tabular}
\end{center}
\caption{Quantitative hallucination comparison of different training approaches.}
\label{tab3}
\end{table}
\begin{table}[t]
\begin{center}
\begin{tabular}{|l|c|c|c|c|c|c|c|c|}
\hline
Method & Bicubic & Ma \emph{et al.} & LapSRN  & UR-DGN & UR-DGN* & SICNN\\
\hline
Identity Similarity & 0.2913 & 0.3823 & 0.4361  & 0.3682 & 0.5267 & \bf{0.5978}\\
\hline
LFW Acc. & 97.51\% & 97.58\% & 97.46\% & 97.20\% & 98.01\% & \bf{98.25}\%\\
\hline
YTF Acc.& 93.08\% & 93.26\% & 93.10\% & 92.78\% & 93.54\% & \bf{93.82}\%\\
\hline
\end{tabular}
\end{center}
\caption{Quantitative comparison on identity recovery and identity recognizability evaluation. The results demonstrate the superiority of our proposed method.}
\label{tab4}
\end{table}

\subsection{Evaluation on Higher Input Resolution}
\begin{figure}
\begin{center}
 \includegraphics[width=1\linewidth]{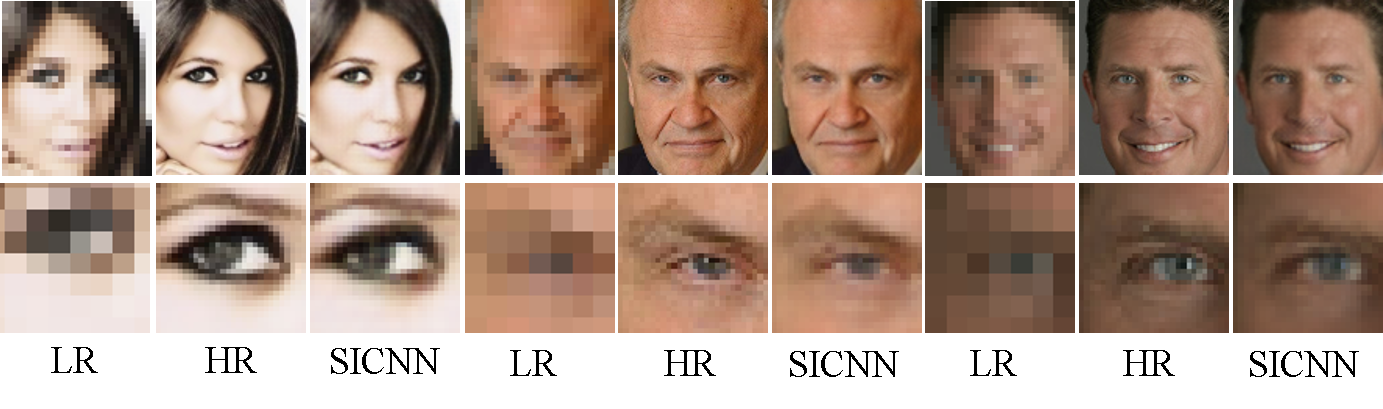}
\end{center}
   \caption{Hallucination visual results for 24$\times$28 inputs with 4$\times$ upscaling factor. Please \emph{zoom in} for better comparison.}
\label{fig:long}
\label{fig:onecol}
\label{figinput}
\end{figure}

For more comprehensive analysis, in this section, we trained our model for 24$\times$28 inputs with 4$\times$ upscaling factor. Specifically, we modify the hallucination network (i.e., $CNN_{H}$) by removing the first DB, DeConv and Conv layers. As shown in Fig. \ref{figinput}, our method performs very well visual quality in higher resolution inputs with 4x upscaling factor.

For identity recovery and identity recognizability evaluation, our method also achieves very good results: Average identity similarity: 0.8868, LFW accuracy: 99.21\%, YTF accuracy: 94.86\%, which are very close to the performance on HR faces.

\subsection{Evaluation on Identity Recovery}\label{sec54}
We perform an evaluation on identity recovery with other state-of-the-art methods. All models for evaluation are the same as last experiment (i.e. Sec. \ref{sec53}).

From the Tab. \ref{tab4}, we observe that our method achieves the best performance. Besides, we also observe that UR-DGN, trained by pixels-wise loss and adversarial loss, even shows inferior performance than LapSRN though with sharper visual results (See Sec. \ref{sec53}). It means that UR-DGN will lose some identity information while super-resolving a face because the adversarial loss is not a pair-wise loss. And if add perceptual loss (i.e. UR-DGN*), pair-wise semantic loss, the results can be improved, but still inferior to our method.

\subsection{Evaluation on Identity Recognizability}\label{sec55}
Follow last two experiments (i.e. Sec. \ref{sec53}, \ref{sec54})., we further perform an evaluation on identity recognizability with other state-of-the-art methods.

From the Tab. \ref{tab4}, we observe that our method achieves the best performance. We also obtain similar observations as last experiment. Besides, we also observe that though several methods (LapSRN. Ma \emph{et al.}, and UR-DGN) obtain better visual results than Bicubic interpolation, the identity recognizability of super-resolved face is similar or even inferior. It means that these methods cannot generate discriminative faces with better identity recognizability.
\subsection{Evaluation on Low-Resolution Face Recognition}\label{sec56}
To evaluate the benefit of our method for low-resolution face recognition, we compare our method ($SICNN+CNN_{R}$) with other state-of-the-art recognition methods on LFW \cite{lfw} and YTF \cite{ytf} benchmark.

From the results in Table \ref{tab6}, we find that these methods' input sizes are relatively large (area size from 15.3$\times$ to 298$\times$ compared with our method). Moreover, using our face hallucination method, the recognition model can still achieve reasonable results in such ultra-low resolution. We also tried using un-aligned faces in training and testing and our proposed method still can achieve similar improvement of performance.

\begin{table}[t]
\begin{center}
\begin{tabular}{|l|c|c|c|c|c|c|c|c|c|}
\hline

Method &  Ours& $CNN_{R}$  & Human& \cite{deepface}& \cite{deepid2+}& \cite{facenet}& \cite{vggface}& \cite{centerloss}&\cite{sphere}\\
\hline
Input Size  & \bf{12$\times$14}& 96$\times$112 & Original & 152$\times$152 & 47$\times$55 & 224$\times$224 &224$\times$224 &96$\times$112 & 96$\times$112 \\
\hline
LFW Acc. & 98.25\% & 99.48\% & 97.53\% & 97.35\% & 98.70\% & 99.63\% & 98.95\% &99.28\% & 99.42\% \\
\hline
YTF Acc.& 93.82\% & 95.38\% & - & 91.4\% & 93.2\% & 95.1\% & 97.3\% &94.9\% & 95.0\% \\
\hline
\end{tabular}
\end{center}
\caption{Face verification performance of different methods on LFW \cite{lfw} and YTF \cite{ytf} benchmark. It shows that our method can help the recognition model to archive high accuracy with ultra-low-resolution inputs.}
\label{tab6}
\end{table}

\section{Conclusion}
In this paper, we present Super-Identity CNN (SICNN) to enhance the identity information during super resolving face images of size 12$\times$14 pixels with an 8$\times$ upscaling factor. Specifically, SICNN aims to minimize the identity difference between the hallucinated face and its corresponding HR face. In addition, we propose a domain-integrated training approach to overcome the dynamic domain divergence problem when training SICNN. Extensive experiments demonstrate that SICNN not only achieves superior hallucination results but also significantly improves the performance of low-resolution face recognition.

\section{Acknowledgement}
This work was supported in part by MediaTek Inc and the Ministry of Science and Technology, Taiwan, under Grant MOST 107-2634-F-002 -007. We also benefit from the grants from NVIDIA and the NVIDIA DGX-1 AI Supercomputer.
%
%
%
%
\bibliographystyle{splncs04}
\bibliography{egbib}
\end{document}